\documentclass[letterpaper, 10 pt, conference]{ieeeconf}  
\usepackage{amssymb}
\IEEEoverridecommandlockouts 

\usepackage{amsmath, amsfonts, amssymb, mathtools, bm}
\usepackage{graphicx}
\usepackage{hyperref}
\usepackage{algorithm}
\usepackage[noend]{algpseudocode}
\usepackage{algorithmicx}
\usepackage{xcolor}
\usepackage{url}
\usepackage{float}
\usepackage{units}
\usepackage{cite}
\usepackage{flushend}
\usepackage[skip=0pt,font=small,labelfont=bf]{caption} % Required for proper captioning
\usepackage{etoolbox}

\definecolor{D}{RGB}{196, 114, 0}
\definecolor{Q}{RGB}{18, 113, 148}

\definecolor{purple}{RGB}{210, 0, 210}
\definecolor{international_orange}{RGB}{240, 74, 0}

\newcommand{\DeformerNet}{\textit{DeformerNet}}

\newcommand{\pcloud}{\mathcal{P}}
\newcommand{\curpcloud}{\pcloud_\mathrm{c}}
\newcommand{\goalpcloud}{\pcloud_\mathrm{g}}

\newcommand{\action}{\mathcal{A}}

\newcommand{\se}{\mathcal{SE}}

\newcommand{\ggn}{\textit{DefGoalNet}}
\newcommand{\diffdef}{\textit{DiffDef}}
\newcommand{\task}{\mathrm{\tau}}
\newcommand{\taskpcloud}{\mathcal{P}_{\mathrm{\tau}}}

\newcommand{\demodata}{\mathcal{D}}

\title{DiffDef: A Diffusion Model for Generating Multimodal Goal Shapes \\ From Demonstrations for Deformable Object Manipulation}
\author{Bao Thach\(^{1,\mathsection}\), Tanner Watts\(^{3,\mathsection}\), Siyeon Kim\(^1\), Britton Jordan\(^1\), Mohanraj Devendran Shanthi\(^1\), Shing-Hei Ho\(^1\), \\James M. Ferguson\(^1\), Tucker Hermans\(^{1,2}\), and Alan Kuntz\(^3\)
\thanks{$\mathsection$ These authors contributed equally; {\tt bao.thach@utah.edu}. $^{1}$Kahlert School of Computing, University of Utah, Salt Lake City, UT, USA; $^{2}$NVIDIA Corporation, Seattle, WA, USA; $^{3}$Department of Electrical and Computer Engineering and Department of Computer Science, Vanderbilt University, Nashville, TN, USA. Research reported in this publication was supported by the Advanced Research Projects Agency for Health (ARPA-H) under the ALISS project, Award Number D24AC00415-00. The ARPA-H award of up to \$11,935,038 provided 100\% of the financial support for this work. The opinions and findings in this paper is solely the responsibility of the authors and do not necessarily represent the official views of ARPA-H.}}

\begin{document}

\maketitle

\begin{abstract}

Deformable object manipulation is a key capability in many robotic applications. 
A promising paradigm for this problem is shape servoing, which aims to control deformable objects toward desired goal shapes. 
However, existing approaches typically rely on impractical goal-shape acquisition methods, such as domain-knowledge engineering or manual manipulation.
Moreover, prior methods generally assume a single deterministic goal and fail to handle multimodal goal settings, a common scenario in many real-world tasks where multiple distinct goal shapes can all lead to successful task completion.
In this paper, we introduce \diffdef{}, a novel neural network that uses a diffusion model to learn a distribution of feasible goal shapes rather than predicting a single deterministic outcome. 
This allows \diffdef{} to generate diverse goal configurations while avoiding the mode-averaging artifacts common in deterministic predictors.
We evaluate our method on several deformable manipulation tasks inspired by manufacturing and surgical applications, both in simulation and on two physical robotic platforms: the da Vinci Research Kit (dVRK) and a bimanual KUKA-based robotic system. 
The results demonstrate that \diffdef{} effectively captures multimodal goal distributions and significantly improves task performance in practical robotic settings.
Website: \url{sites.google.com/view/diffdef}.

\end{abstract}

%%%%%%%%%%%%%%%%%%%%%%%%%%%%%%%%%%%%%%%%%%%%%%%%%%%%%%%%%%%%%%%%%%%%%%%%%%%%%%%%
\section{Introduction}\label{sec:intro}
Deformable object manipulation is a central problem in robotics, with applications in surgery, manufacturing, and everyday household environments.
Unlike rigid bodies, deformable objects exhibit effectively infinite degrees of freedom and complex dynamics~\cite{Sanchez2018robotic, zhu2022challenges, Wu2020Learning}, making it difficult to explicitly define task goals or learn policies that generalize across contexts.
A key limitation of prior approaches~\cite{Navarro-Alarcon2016, qi2022model, alambeigi2018robust, alambeigi2018autonomous, zhu2021vision, shetab2022lattice} is their reliance on manually specified geometric goal shapes~\cite{thach2024defgoalnet}.
The recent state-of-the-art method, \ggn{}~\cite{thach2024defgoalnet}, addresses this limitation by learning to generate goal shapes directly from human demonstrations.

\begin{figure}
    \centering
    \includegraphics[width=\linewidth]{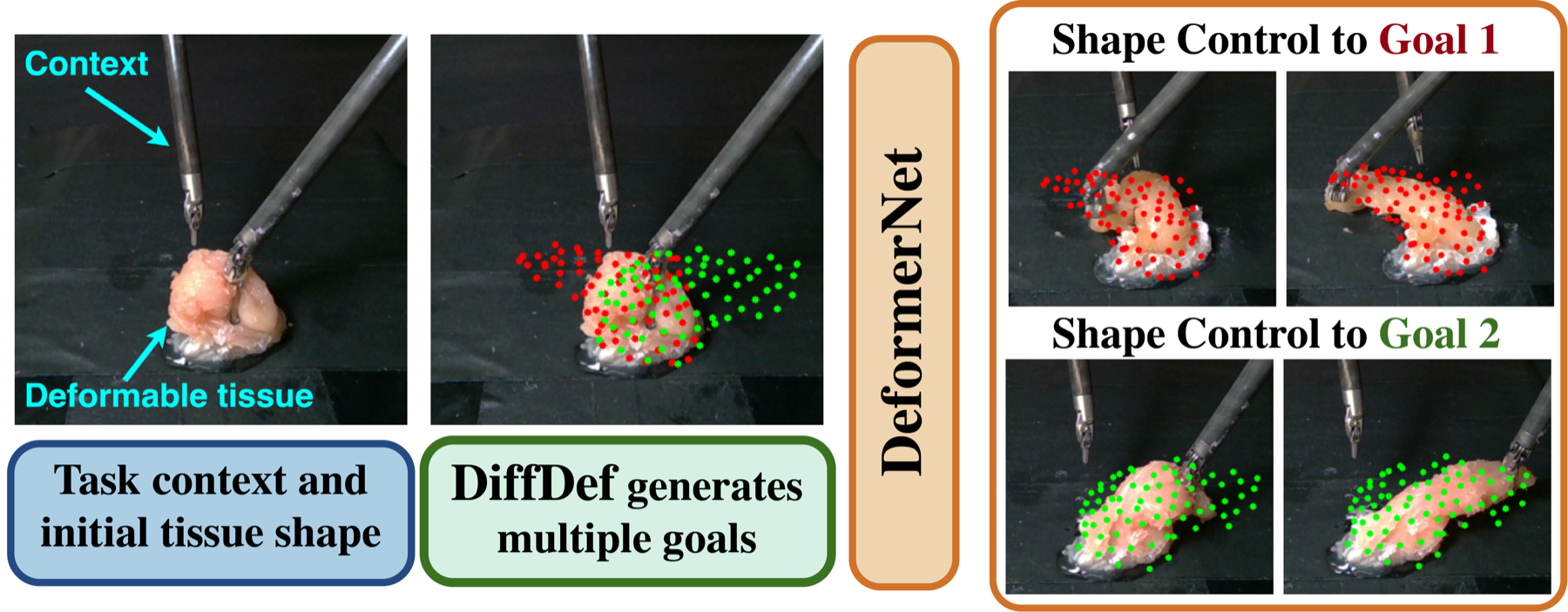}
    \vspace{-5pt}
    \caption{\textbf{Multimodal Contextual Shape Servoing:} Given the task context and initial shape of the deformable object, \diffdef{} predicts multiple valid goal point clouds (red and green points) by learning a diffusion-based probabilistic model from human demonstrations. 
    Conditioned on one of these predicted goals, \DeformerNet{} then computes actions to manipulate the object into the desired shape and successfully accomplishes the task. We experiment on a surgical retraction task, where the robot is tasked with pulling the tissue aside to create tension without colliding with the surgical tool.} 
    \label{fig:diffdef_overview}
\end{figure}

However, \ggn{} predicts only a single deterministic goal shape and struggles to represent multimodal goal distributions. Such multimodal settings are common and critical in real-world robotic tasks.
Multimodality occurs when multiple distinct shapes can equally accomplish a task under the same conditions.
For example, a surgical robot may deform tissue either to the left or to the right, both resulting in successful outcomes in a tissue retraction task (Figure~\ref{fig:diffdef_overview}).
In such cases, deterministic models like \ggn{} ``average'' these possibilities (Figure~\ref{fig:diffdef_vs_defgoalnet_retraction_tool_sim}), producing an ambiguous and physically infeasible goal shape that often leads to task failure.

To address this limitation, we introduce \diffdef{}, a novel neural network that models multimodal goal distributions by generating goal shapes conditioned on geometric context via a conditional diffusion process~\cite{ho2020denoising}.
As part of our learning-from-demonstration shape servoing pipeline, \diffdef{} is trained on demonstration trajectories collected from diverse policies (e.g., different human demonstrators performing the same task in distinct ways). 
This enables the model to capture the full distribution of goal shapes that successfully accomplish a task given the current context.
At runtime, \diffdef{} takes as inputs the current deformable object point cloud and a contextual point cloud, and generates a goal shape sampled from the learned distribution. 
The predicted goal is then passed to \DeformerNet{}~\cite{thach2023deformernet, thach2022learning}, which computes robot actions to deform the object toward the goal shape, thereby successfully completing the
task.

We evaluate our approach in simulation and in real-world experiments on both surgical (tissue retraction) and manufacturing (object packaging) tasks. 
Across all tasks, the multimodal goal shapes generated by \diffdef{} consistently achieve high success rates, outperforming the state-of-the-art goal generation method \ggn{}. 
By introducing a generative, multimodal formulation for deformable object goals, our approach brings robotic manipulation of soft materials closer to practical real-world applications.

\section{Related Work}\label{sec:related_work}
Recent advances in machine learning have empowered robots to manipulate rigid objects by effectively leveraging rich, high-dimensional sensory inputs such as 3D point clouds~\cite{mousavian20196, murali20206, deng2020self, lu2020multi, van2020learning}. 
Neural networks, in particular, have become central to tackling complex robotic perception and control problems, including shape reconstruction~\cite{van2020learning}, object pose estimation~\cite{lu2020multi}, and grasp synthesis~\cite{mousavian20196, lu2020multi, van2020learning}. 
Furthermore, sophisticated learning frameworks have enabled robots to execute long-horizon, multistep tasks by integrating a diverse set of learned behaviors~\cite{brohan2022rt}. Building on these successes, we explore a learning-based framework aimed at controlling the 3D shape of deformable objects.

Traditionally, shape servoing has been addressed through predominantly model-based or learning-free techniques~\cite{alambeigi2018autonomous, navarro2016automatic, qi2022model, alambeigi2018robust, shetab2022lattice}. 
These methods typically rely on manually selected feature points to represent the deformable object, making them susceptible to sensor noise and limiting their ability to generalize to novel object geometries.
Shetab-Bushehri \emph{et al.}~\cite{shetab2022lattice} propose a 3D lattice-based representation that enables accurate 3D control of deformable shapes. However, their method relies on the assumption of consistent feature correspondences over time, which is difficult to uphold in real-world robotic deployments.

Learning-based approaches have emerged as a promising direction for 3D shape control by leveraging the generalization strength of modern neural networks. Hu \emph{et al.}~\cite{Hu20193-D} utilize Fast Point Feature Histograms~\cite{Rusu2010VFH} to represent the state of deformable objects within a learning framework. However, this architecture struggles to model the complex dynamics of 3D deformable objects~\cite{thach2022learning}. The current state-of-the-art method, \DeformerNet{}~\cite{thach2022learning, thach2023deformernet}, addresses this issue by employing a PointConv-based~\cite{wu2018pointconv} neural network that inputs the current and goal point clouds and predicts the robot action to deform the object toward the goal shape. 
However, all of these methods rely on explicitly defined goal shapes, a key limitation precluding real-world applications.

Although point cloud generative models have achieved impressive results in computer vision and graphics~\cite{shu20193d, achlioptas2018learning, arshad2020progressive, pumarola2020c}, their integration into robotic applications has so far been relatively underexplored.

Recent methods in robotic goal generation synthesize images from language instructions~\cite{van2022large, du2023learning}, but 2D goals cannot capture the complex geometries required for deformable object control. Text-only goals~\cite{mees2022matters, zhou2023bridging} also fall short, as they lack the geometric details needed for 3D shape servoing.

For learning robotic policies from multimodal demonstration datasets, diffusion-based models such as Diffusion Policy~\cite{chi2023diffusion} and Stein Variational Belief Propagation~\cite{pavlasek2024stein} directly map observations to actions, yet they demand costly demonstrations and large-scale datasets. Our approach instead predicts multimodal goal representations, leaving action execution to \DeformerNet{}. This separation improves sample efficiency: action data can be generated at low cost in simulation, while expensive demonstration data is required in much smaller quantities for goal learning~\cite{thach2024defgoalnet}.

\section{Problem Formulation}\label{sec:problem}
We study robotic manipulation tasks involving deformable objects where success can be achieved through multiple distinct strategies. For example, in tissue retraction, pulling tissue either to the left or right may both generate the necessary exposure of underlying anatomy. In such settings, the robot must reason not about a single goal state, but about a distribution of feasible goal shapes.

Formally, let $\curpcloud$ denote the current partial-view point cloud of the deformable object, and let $\taskpcloud$ represent a contextual point cloud capturing task-relevant environmental features. Together, $(\curpcloud, \taskpcloud)$ define the scene state. Given a task $\task$, our objective is to model a conditional distribution
\begin{equation}
    p(\goalpcloud \mid \curpcloud, \taskpcloud),
    \label{eq:prob_form}
\end{equation}
where $\goalpcloud$ is a feasible goal point cloud that would lead to successful task completion of $\task$.

\begin{figure}
    \centering
    \includegraphics[width=1.0\linewidth]{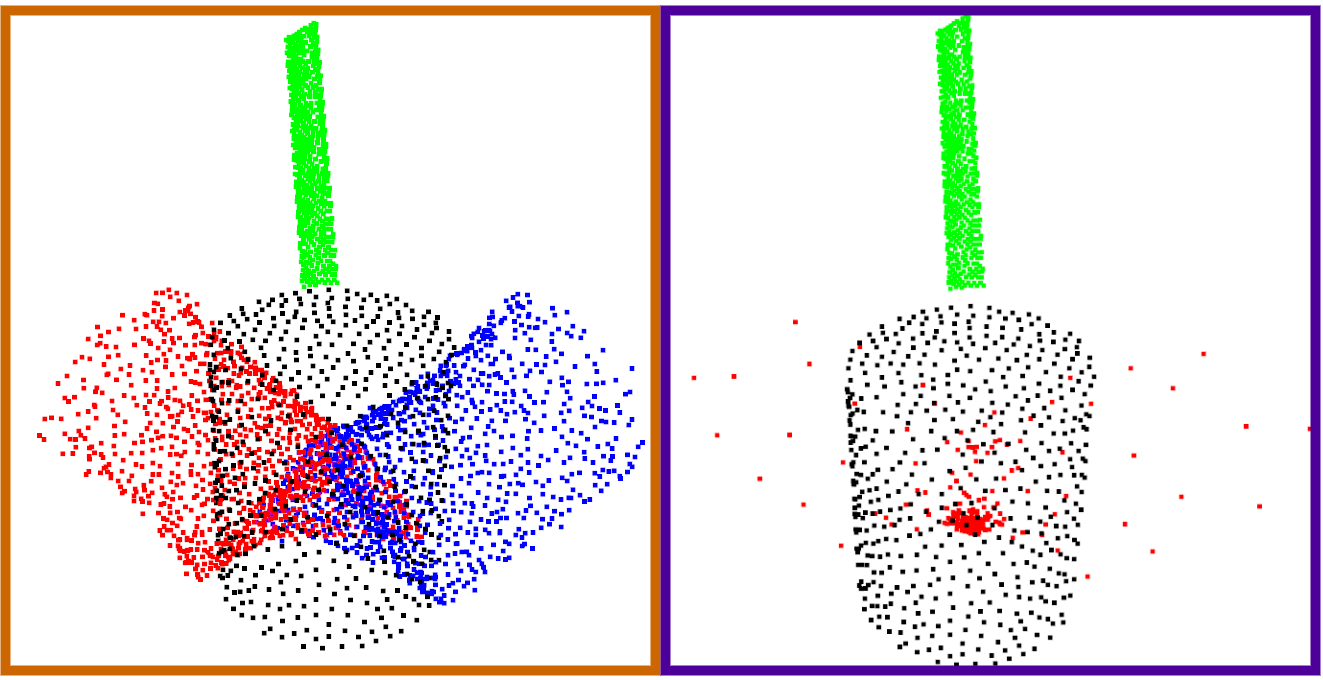} 
    \vspace{-7pt}
    \caption{Simulated retraction results comparing predicted goals from \diffdef{} (left) to \ggn{} (right). \diffdef{} (this paper) effectively captures the underlying bimodal goal distribution, producing two distinct, realistic goal shapes (red and blue).
    In contrast, \ggn{} (current state-of-the-art) averages the two modes, resulting in a single unrealistic and physically infeasible goal point cloud (red).} 
    % \jmf{consider separating the boxes here just a little bit. Red/blue seem to overlap.} 
    \label{fig:diffdef_vs_defgoalnet_retraction_tool_sim}
\end{figure}

We assume access to a multimodal expert demonstration dataset $\demodata$, where each trajectory contains sequences of object and contextual point clouds illustrating different but equally valid strategies for completing $\task$. The multimodality of $\demodata$ ensures that $p(\goalpcloud \mid \curpcloud, \taskpcloud)$ reflects the diversity of successful outcomes rather than collapsing them into a single average goal. 

At execution time, the robot samples $\goalpcloud \sim p(\goalpcloud \mid \curpcloud, \taskpcloud)$ and uses a downstream policy (\DeformerNet{}~\cite{thach2023deformernet, thach2022learning}) to compute the actions that deform the object toward the sampled goal. 
We formulate robot action $\action$ as rigid transformation $\se(3)$, representing the change of robot end-effector's pose.
% This separation of goal generation from action execution provides flexibility through multi-modal goal sampling and efficiency since task-agnostic action supervision can be collected at scale in simulation as in~\cite{thach2021deformernet}.  

\section{Methods}\label{sec:method}
Our framework decomposes deformable object manipulation into two subproblems: (i) \textbf{goal generation}, where the robot predicts a distribution over feasible goal shapes given the current state and task context, formalized as $p_\theta(\goalpcloud \mid \curpcloud, \taskpcloud)$; and (ii) \textbf{goal-conditioned control}, where the robot executes actions to deform the object toward a sampled goal shape. 
This separation improves sample efficiency: expensive demonstrations are required only for goal learning, while control policy supervision can be collected at scale in simulation.  
Since \DeformerNet{}~\cite{thach2023deformernet} already provides a robust solution for goal-conditioned control, our focus in this section is on solving the goal learning problem via \diffdef{}. 

\subsection{Conditional Goal Distribution Learning}

We represent the set of all valid outcomes for a task as a conditional distribution over goal point clouds as in Equation~\ref{eq:prob_form}, where $\curpcloud$ is the current partial-view object point cloud and $\taskpcloud$ encodes task-relevant context. 
Unlike prior deterministic goal networks~\cite{thach2024defgoalnet}, this formulation captures the inherent multimodality of deformable manipulation tasks.  

To parameterize $p_\theta(\goalpcloud \mid \curpcloud, \taskpcloud)$, we adopt a \textit{conditional diffusion model}~\cite{ho2020denoising}. 
Given a dataset $\demodata$ of demonstrations with diverse, successful strategies, we add noise to each goal point cloud $\goalpcloud$ through the forward diffusion process, and train a noise predictor $\mu_\theta$ to recover the injected noise conditioned on $(\curpcloud, \taskpcloud)$. 
At inference, we generate diverse samples $\goalpcloud \sim p_\theta$ by reversing this diffusion process.

\subsection{\diffdef{} Architecture}

\begin{figure}
    \includegraphics[width=\linewidth, trim=30pt 0pt 0pt 0pt, clip]{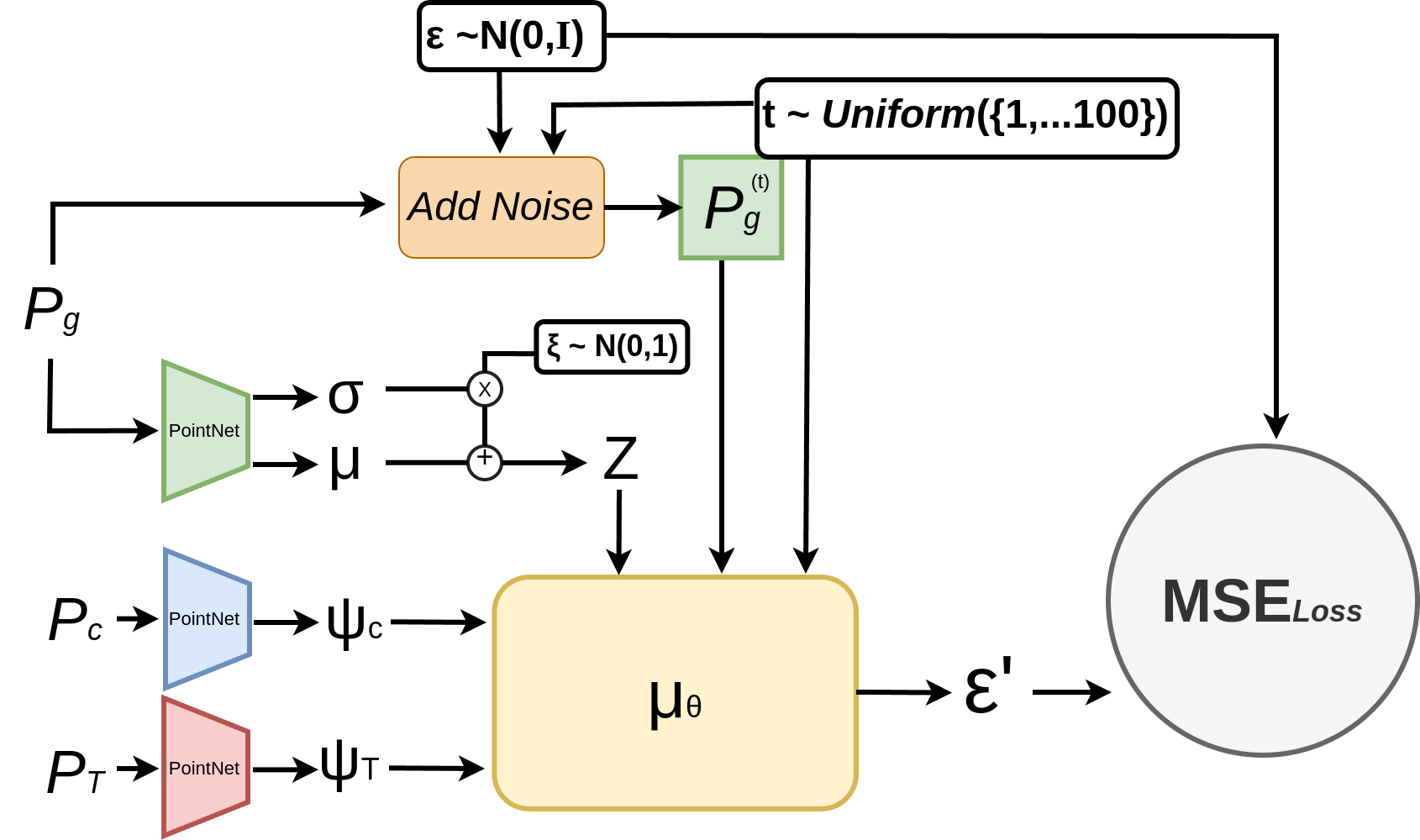}
    \vspace{-5pt}
    \caption{\diffdef{} architecture for learning multimodal goal distributions.  
    $P_g$: goal point cloud; $P_c$: current point cloud; $P_T$: context. Encoders produce latent representations $(z, \psi_c, \psi_T)$.  
    The noise predictor $\mu_\theta$ estimates $\epsilon'$ to match ground-truth noise $\epsilon$.}    
    \label{fig:diffdef_architecture}
\end{figure}

Figure~\ref{fig:diffdef_architecture} summarizes our model architecture.  
A PointNet~\cite{pointnet} encoder maps the goal point cloud $\goalpcloud$ into Gaussian parameters $(\mu, \sigma)$, from which we sample a latent vector $z$. 
Two additional PointNet encoders map the current and contextual point clouds $(\curpcloud, \taskpcloud)$ into feature vectors $(\psi_c, \psi_\task)$. 
A noisy version of the goal $\goalpcloud^{(t)}$ is obtained via the forward diffusion process~\cite{ho2020denoising, luo2021diffusion} for training supervision, where $t$ denotes the diffusion timestep. 
Finally, the \textit{noise predictor} $\mu_\theta$ takes as input $(\goalpcloud^{(t)}, t, z, \psi_c, \psi_\task)$ and outputs the predicted noise $\epsilon'$. 
The model is trained to minimize the reconstruction loss between $\epsilon'$ and the ground-truth noise $\epsilon$, with an additional Kullback–Leibler (KL) regularization term encouraging a smooth latent space.

The forward diffusion process~\cite{ho2020denoising, luo2021diffusion} injects noise into a goal point cloud $\goalpcloud$ over multiple time steps $t \sim \mathcal{U}\{1,\dots,T\}$ to produce a noisy version $\goalpcloud^{(t)}$:
\begin{equation}
    \goalpcloud^{(t)} = \sqrt{\bar{\alpha}_t} \goalpcloud + \sqrt{1 - \bar{\alpha}_t} \, \epsilon, 
    \quad \epsilon \sim \mathcal{N}(0, I),
\end{equation}
where $\bar{\alpha}_t$ is the cumulative noise schedule.  
In our framework, we use a PointNet-based goal encoder to produce a latent sample
\begin{equation}
    z = \mu + \sigma \odot \xi, \quad \xi \sim \mathcal{N}(0, I).
\end{equation}
Given the point clouds, the current and contextual encoders yield
\begin{equation}
    \psi_c = \text{Enc}(\curpcloud), \quad \psi_\task = \text{Enc}(\taskpcloud).
\end{equation}
The noise predictor model then estimates the added noise from all of the encoded information:
\begin{equation}
    \epsilon' = \mu_\theta(\goalpcloud^{(t)}, t, z, \psi_c, \psi_\task).
\end{equation}

The training objective is the mean squared error between predicted and ground-truth noise, regularized by a KL penalty on the latent space:
\begin{equation}
    \mathcal{L} = \mathbb{E}\left[ \| \epsilon - \epsilon' \|^2 \right] 
    + \lambda 
D_{\mathrm{KL}}\big[q_\varphi(z \mid \goalpcloud) \,\|\, \mathcal{N}(0,I)\big].
\end{equation}

At runtime, we sample $z \sim \mathcal{N}(0,I)$ and initialize $\goalpcloud^{(T)} \sim \mathcal{N}(0,I)$. 
The reverse diffusion process~\cite{ho2020denoising, luo2021diffusion} iteratively removes noise using $\mu_\theta$, ultimately yielding a clean goal $\hat{\goalpcloud} \sim p_\theta(\goalpcloud \mid \curpcloud, \taskpcloud)$. 
Sampling multiple goals from this distribution produces diverse valid strategies to accomplish the task $\task$.  

\subsection{Integration With Goal-Conditioned Control}

Once a goal point cloud is successfully generated by \diffdef{}, the robot deforms the object to reach the goal shape using \DeformerNet{}~\cite{thach2022learning}, a task-agnostic deformable shape control policy that maps $(\curpcloud, \goalpcloud) \mapsto \action$. 
Executed in a closed-loop manner~\cite{thach2022learning, thach2024defgoalnet, thach2023deformernet}, this policy drives the object toward the sampled goal, thereby successfully completing the manipulation task $\task$. 
Together, \diffdef{} and \DeformerNet{} form a unified pipeline: \diffdef{} proposes diverse, task-feasible goal shapes, while \DeformerNet{} efficiently executes corresponding deformation actions.

\section{Experiments and Results}\label{sec:experiments}
\begin{figure}
    \centering
    \includegraphics[width=1.0\linewidth]{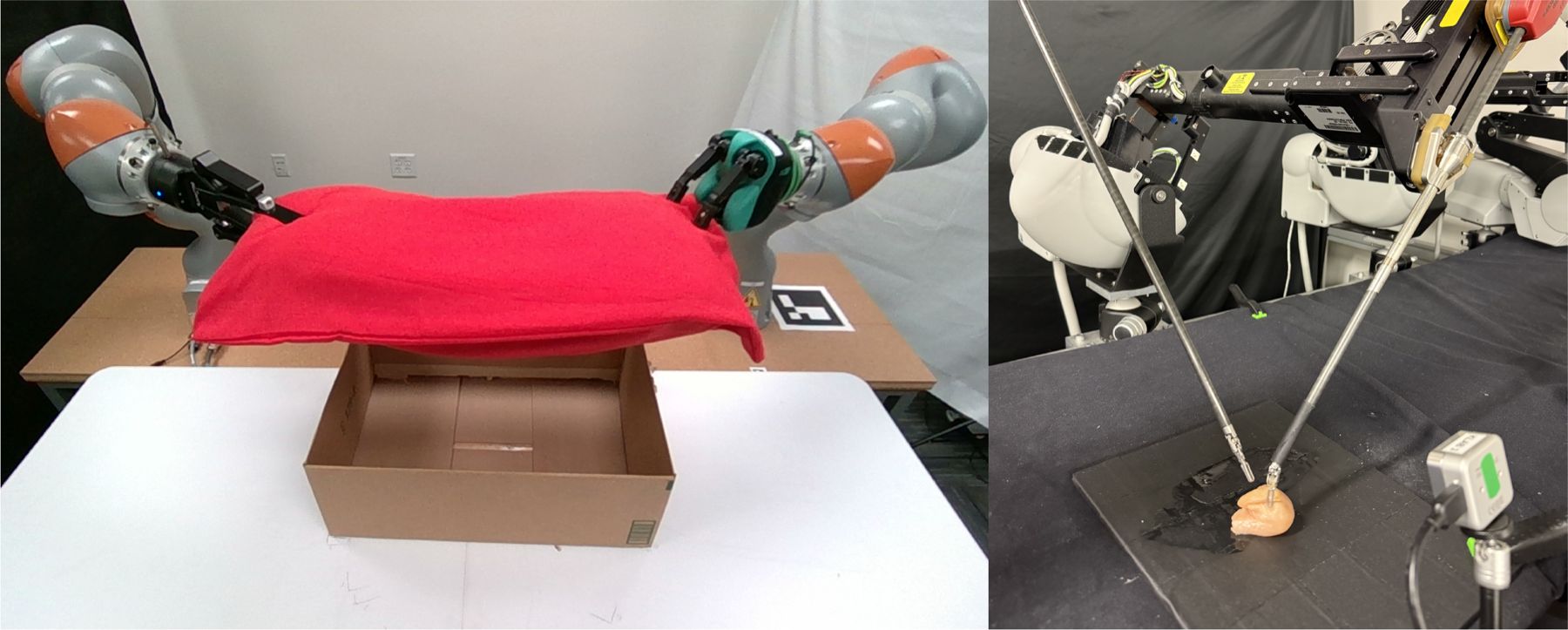} 
    \vspace{-5pt}
    \caption{Physical robot experiment setups. \textbf{(Left)} For the object packaging task, we use two KUKA robotic arms to manipulate a soft pillow. \textbf{(Right)} For the surgical retraction task, we use a dVRK surgical robot to manipulate \textit{ex vivo} chicken muscle tissue.} 
    \label{fig:physical_exp_setup}
\end{figure}

We evaluate the performance of \diffdef{} on two representative robotic tasks that highlight its versatility across both surgical and manufacturing domains: (i) \textbf{tissue retraction}, which captures the safety-critical requirements of soft-tissue surgery, and (ii) \textbf{object packaging}, which reflects deformable object manipulation in industrial automation.  
For each task, we study both simulated environments and physical robot setups.  
This structure enables us to compare controlled, large-scale simulation results with real-world outcomes under practical sensing and actuation conditions. 

In simulation, we utilize the patient-side manipulator of the da Vinci Research Kit (dVRK) surgical robot~\cite{Kazanzides2014_ICRA_DVRK}, using the Isaac Gym platform~\cite{Liang2018GPU}. 
For physical-robot experiments, we use the physical dVRK system for the surgical task, and two KUKA iiwa 7-DOF robotic arms equipped with Robotiq and Reflex Takktile 2 grippers for the manufacturing task.
An Intel\textsuperscript{\textregistered} RealSense\textsuperscript{\texttrademark} D455 camera is used to capture point clouds for perception.
The physical experiment setups are visualized in Figure~\ref{fig:physical_exp_setup}.

\begin{figure}
    \includegraphics[width=1\linewidth]{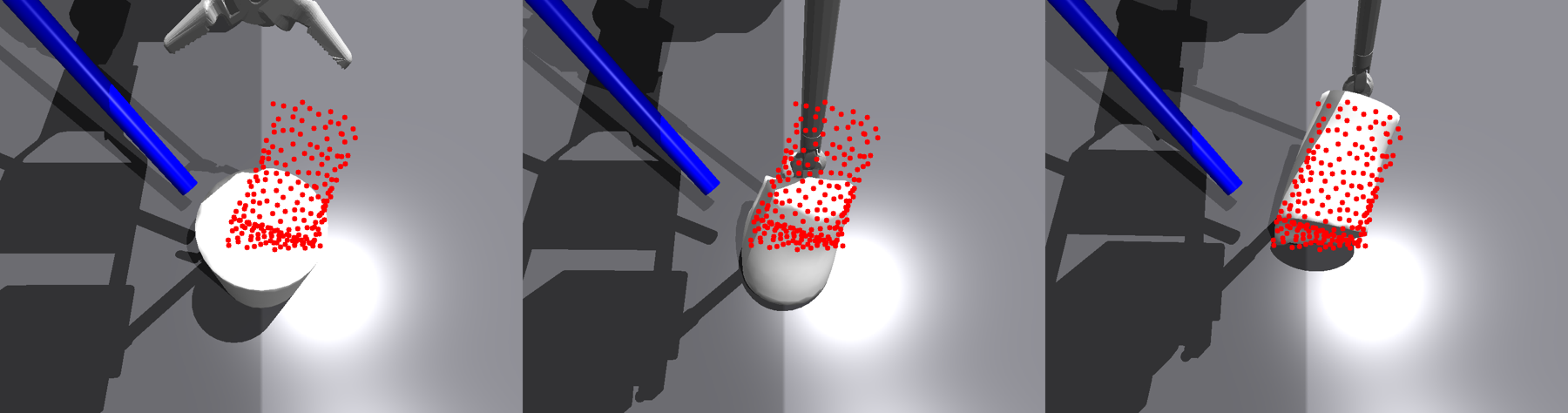}
    \vspace{-6pt}
    \caption{Representative manipulation sequence of retraction task in simulation. Red point cloud is the goal shape generated by \diffdef{}.}
    \label{fig:sequence_retraction_tool_sim}
\end{figure}

\begin{figure}
    \centering
    \vspace{5pt}
    \includegraphics[width=0.7\linewidth]{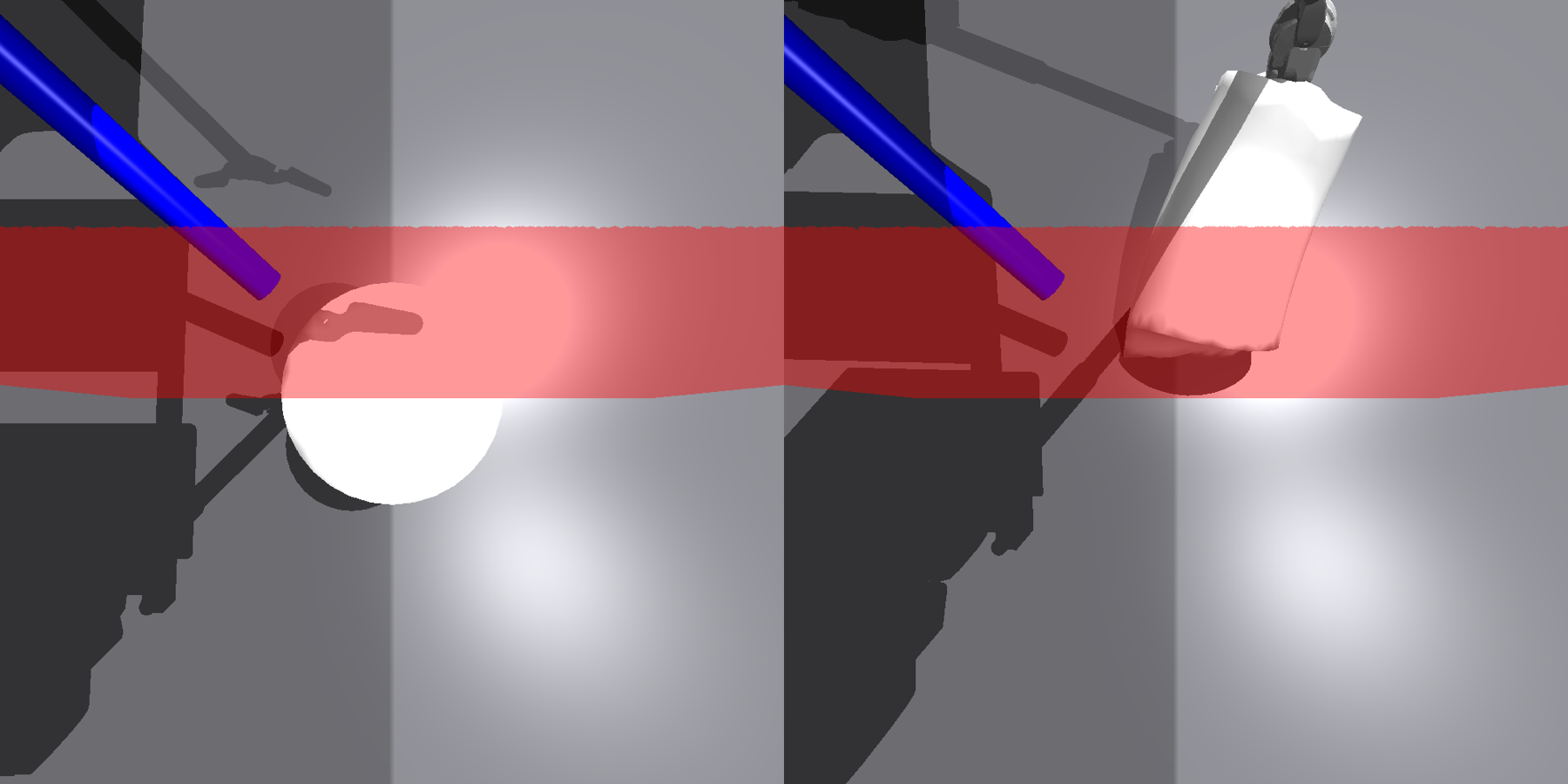}
    \vspace{5pt}
    \caption{Tool-conditioned surgical retraction task setup. The objective is to pull the tissue (white) to one side to create tension for subsequent cutting, while avoiding collision with the other surgical tool (blue). \textbf{Left:} Initial task state, where a target plane (red) bisects the tissue into two halves. \textbf{Right:} The task is deemed successful if the tissue is fully retracted past the target plane without any collision.}
    \label{fig:before_and_after_retraction_plane}
    % \vspace{-20pt}
\end{figure}

\begin{figure}
    \centering
    \vspace{5pt}
    \includegraphics[width=0.325\linewidth]{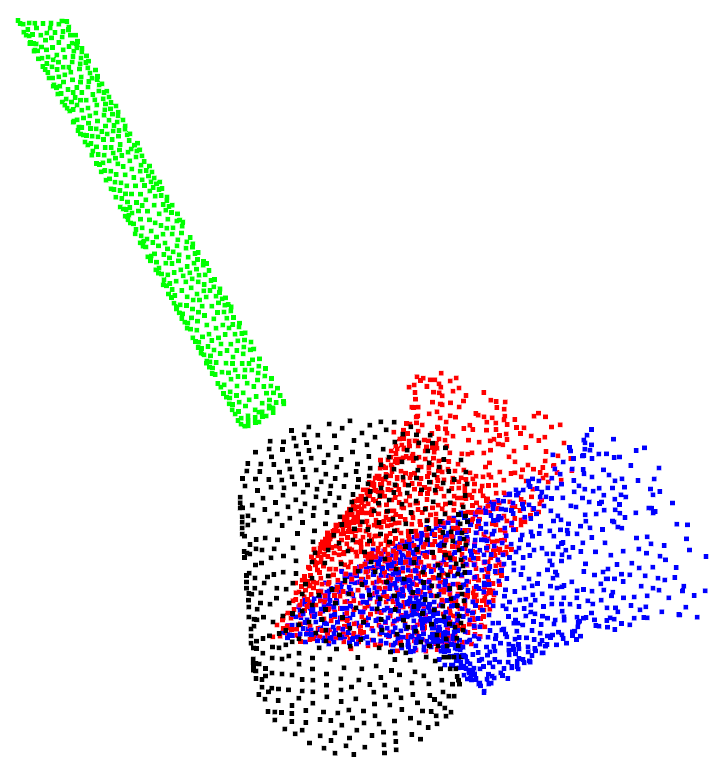} 
    \includegraphics[width=0.325\linewidth]{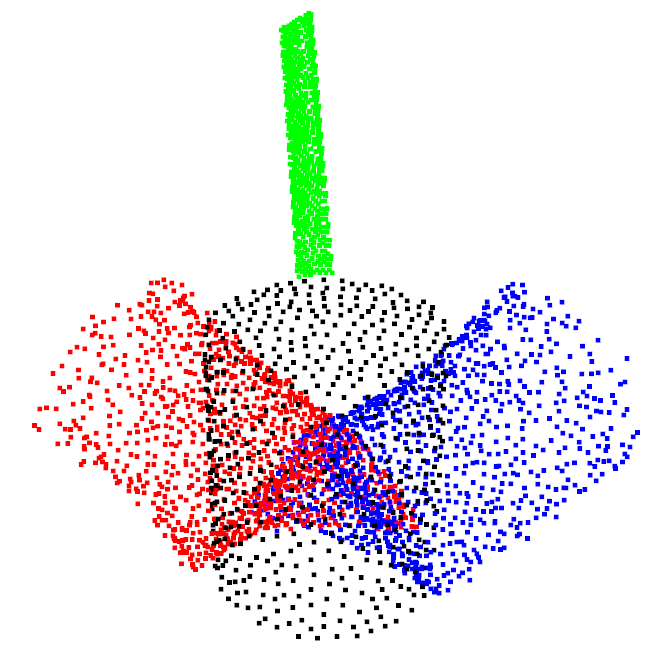}    
    \includegraphics[width=0.325\linewidth]{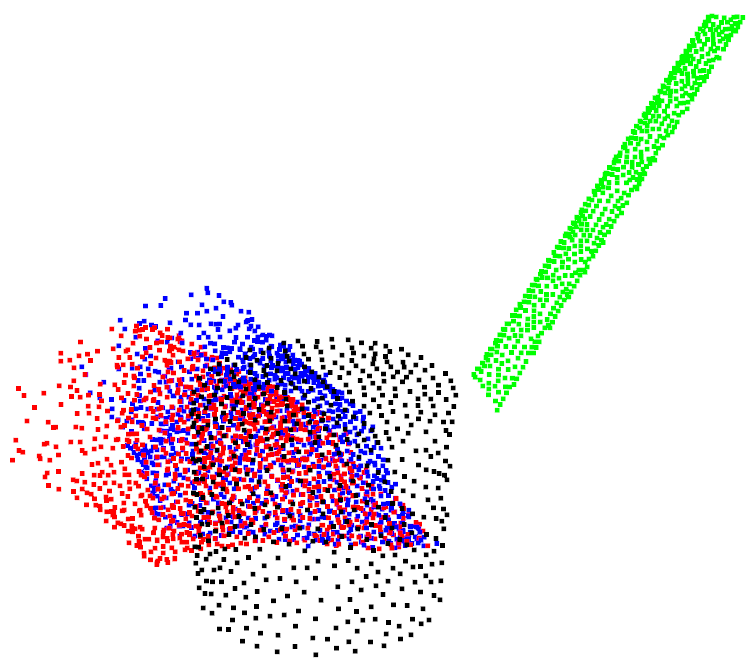} 
    \caption{Example demonstrations for the surgical retraction task. Black: initial tissue state; green: contextual surgical tool; red/blue: two distinct, equally valid goal shapes.}
    \label{fig:demo_example_retraction_tool}
\end{figure}

\subsection{Tissue Retraction}
\label{sec:exp_retraction}

Retraction is a fundamental task in soft-tissue resection: tissue must be placed under sufficient tension to expose the cut target and enable safe resection.  
A key challenge is not just whether to retract, but how to select a retraction direction that provides effective exposure while avoiding collisions with other surgical instruments or sensitive anatomy.    
We design retraction experiments, in both simulation and on physical hardware, to capture this multimodal decision-making process and evaluate the extent to which \diffdef{} can generate diverse feasible retraction goals. 

\subsubsection{Simulated Retraction}
\label{sec:exp_retraction_sim}

We first evaluate \diffdef{} on a simulated tissue retraction task built in Isaac Gym using the dVRK patient-side manipulator (PSM) robot~\cite{Kazanzides2014_ICRA_DVRK}.
The simulated tissue is modeled as a cylindrical deformable patch, and the cutting instrument (to avoid collisions with) is represented by a cylindrical rigid tool inserted into the scene (see Figure~\ref{fig:sequence_retraction_tool_sim}). The contextual point cloud $\taskpcloud$ corresponds to the visible portion of this tool, constraining feasible retraction directions.
The robot must pull the tissue so that it crosses a predefined target plane (representing ``safe exposure'') without colliding with the tool (see Figure~\ref{fig:before_and_after_retraction_plane}).

\paragraph{Dataset}
In simulation, we collect multimodal demonstrations using a scripted multimodal robot policy (Figures~\ref{fig:before_and_after_retraction_plane} and~\ref{fig:demo_example_retraction_tool}), which produces multiple demonstrations for each context. 
For each tool pose, two distinct retraction angles are executed (e.g., $30^\circ$ and $60^\circ$ relative to the tool axis), creating multiple valid goals per context (see Figure~\ref{fig:demo_example_retraction_tool}). 
We use tissues with diameters ranging between 1–7 cm, consistent with the typical distribution of tumor sizes observed in surgery~\cite{ernst2004central}. We randomize tissues poses and tool poses to define restricted regions where tissue retraction is not allowed.
Note that we use simulated data only for extensive ablation studies and to understand the impact of demonstration dataset size on \diffdef{} performance. All subsequent real robot experiments are conducted with models trained on real human demonstrations.

\begin{figure}
    \includegraphics[width=\linewidth]{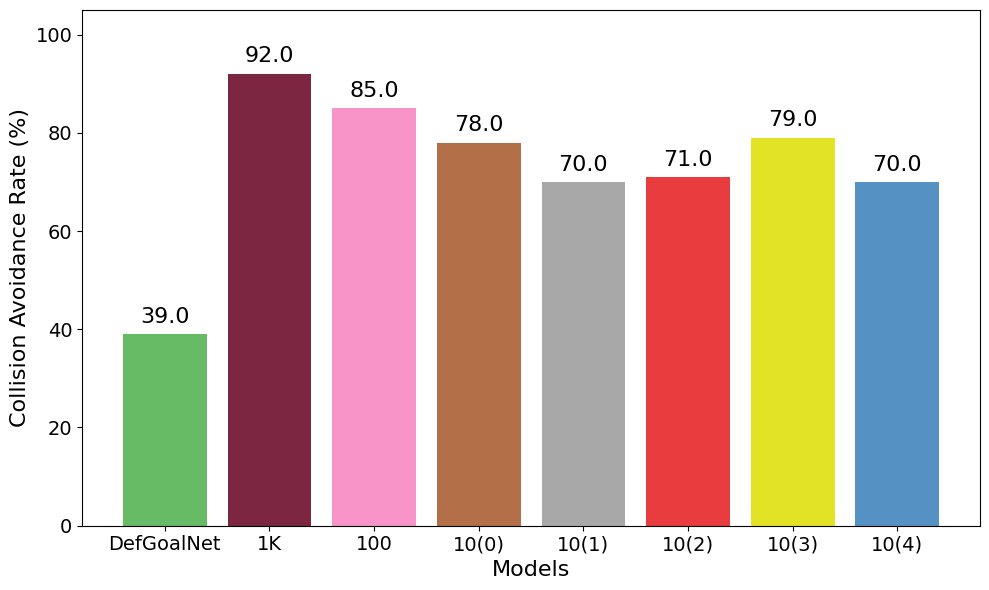}  
    \caption{\textbf{Simulated retraction}---Collision-avoidance rate (higher is better) across multiple dataset sizes. Left to right: \ggn{} trained with 1000 demos vs \diffdef{} with 1000, 100, and 10 demos.}
    \label{fig:retraction_eval_collision_avoid_diffdef_vs_defgoalnet}
\end{figure}

\paragraph{Training and Evaluation}
We train \diffdef{} with demonstration datasets of size 10, 100, and 1000, and evaluate on 100 unseen contexts. Baselines include the deterministic \ggn{}~\cite{thach2024defgoalnet}.  
Metrics include:  
(i) \textit{Collision-avoidance rate}: percentage of executed retraction sequences that successfully avoid tool collision.  
(ii) \textit{Success percentage}: percentage of points in the final tissue point cloud that successfully pass through the target plane. Note that we only calculate this metric for retraction sequences that are \textit{collision-free}.
(iii) \textit{Chamfer distance}: geometric similarity between predicted and ground-truth goal point clouds. 

\begin{figure}
    \includegraphics[width=\linewidth]{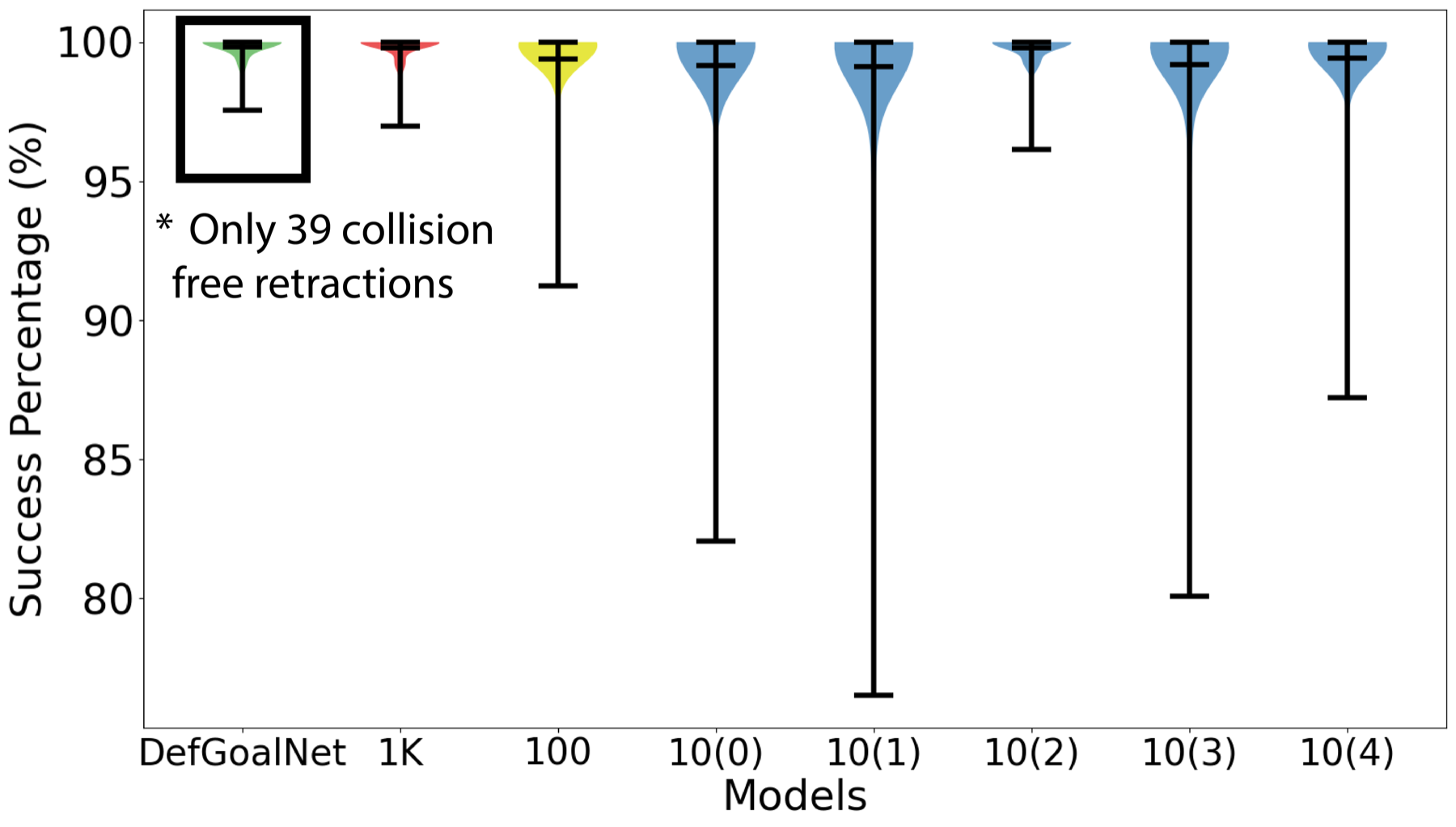} 
    \caption{\textbf{Simulated retraction results}---Success percentage metric. We only calculate this metric on collision-free retraction sequences.}    
    \label{fig:retraction_eval_success_precentage}
\end{figure}

\begin{figure}
    \centering
    \includegraphics[width=\linewidth]{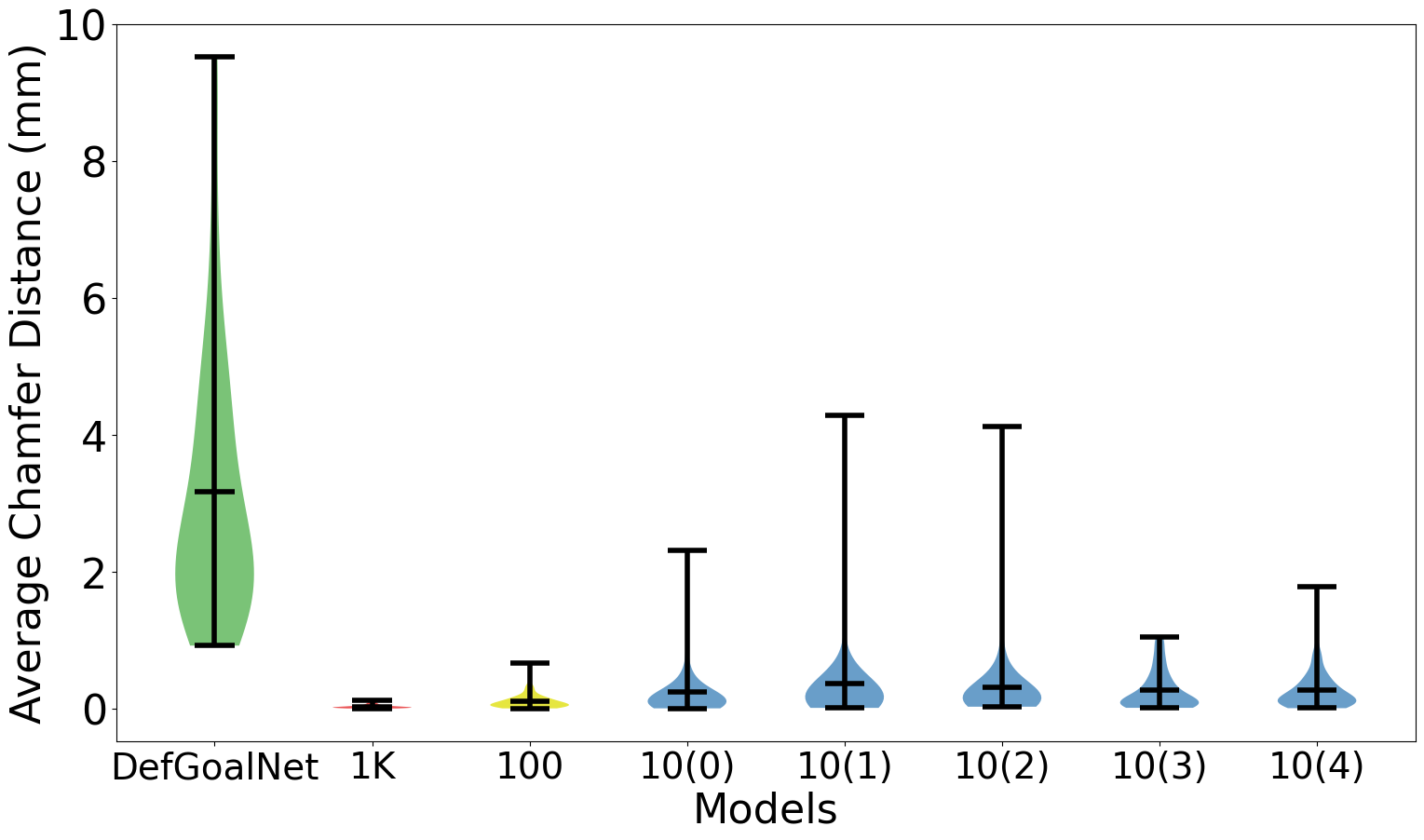}  
    \caption{\textbf{Simulated retraction results}---Violin plots of Chamfer distances (lower is better) between predicted and ground truth goals.}
    \label{fig:retraction_eval_chamfer_diffdef_vs_defgoalnet}
\end{figure}

\paragraph{Results} 
Qualitatively, as illustrated in Figure~\ref{fig:diffdef_vs_defgoalnet_retraction_tool_sim}, \diffdef{} captures much more effectively the underlying two-mode goal distribution, producing two distinct and equally valid goals. 
Both predicted goals are realistic and semantically coherent. 
In contrast, \ggn{} average the two modes, resulting in a single unrealistic and physically impossible goal point cloud.
Quantitatively, even with as few as 10 demonstrations, \diffdef{} achieves a collision avoidance rate of $>$70\% and a median success percentage of $>$95\% (Figures~\ref{fig:retraction_eval_collision_avoid_diffdef_vs_defgoalnet} and \ref{fig:retraction_eval_success_precentage}).
Notably, \diffdef{} trained on just 10 demonstrations already outperforms \ggn{} trained with the full dataset of 1000 demonstrations. For success percentage, \diffdef{} appears comparable to \ggn{}, but only 39 collision-free retractions from \ggn{} are included in the calculation, compared to the 92 retractions from \diffdef{}.
Representative manipulation sequences are shown in Figure~\ref{fig:sequence_retraction_tool_sim}.  
The retraction simulation experiments confirm the key advantages of \diffdef{}: robustness to small datasets, the ability to capture multimodal goal distributions, and superior task success compared to \ggn{}. 

\subsubsection{Physical Retraction}
\label{sec:exp_retraction_real}

\begin{figure}
    \centering
    \includegraphics[width=0.8\linewidth]{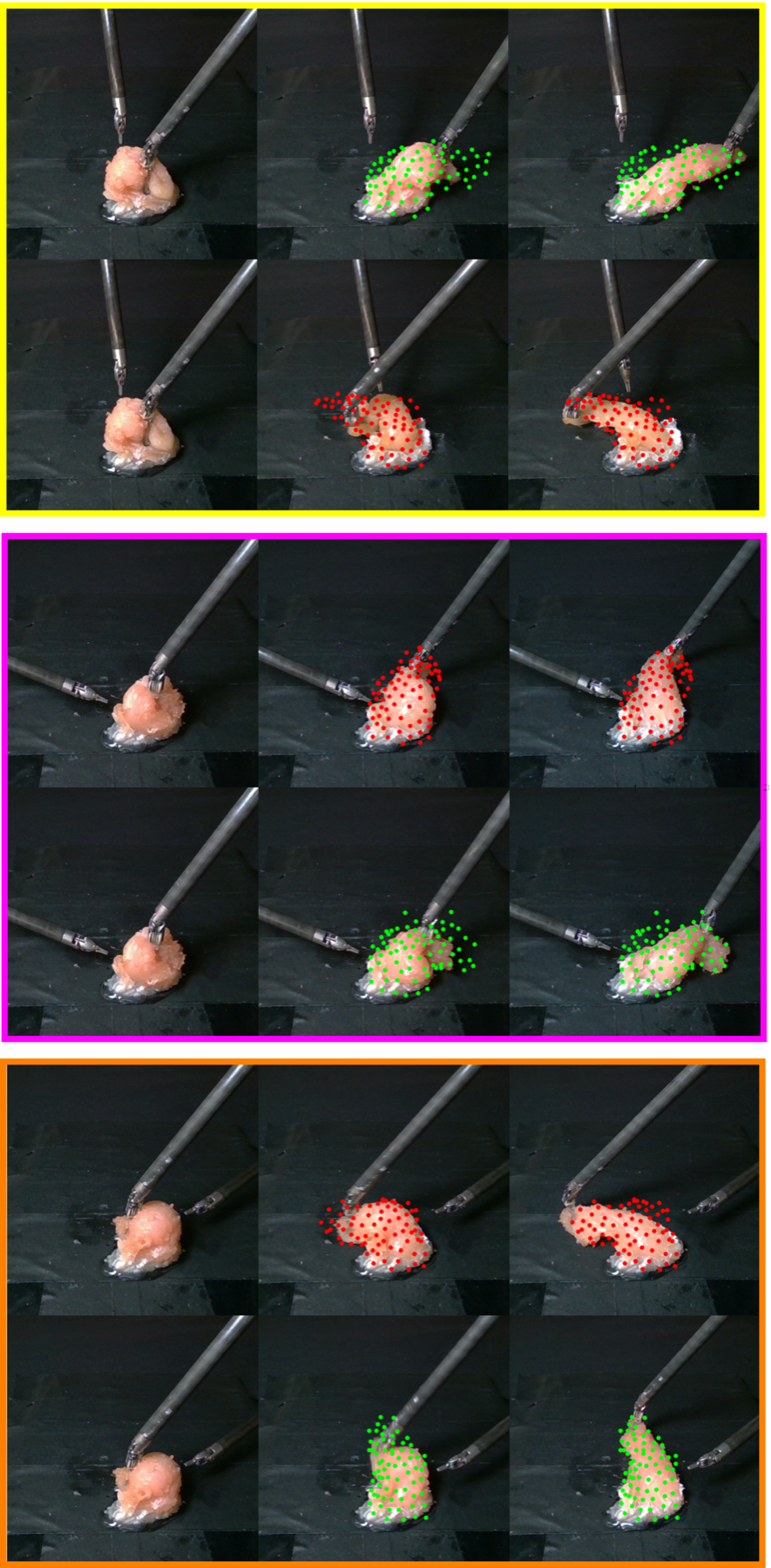}  
    \vspace{3pt}
    \caption{Example retraction sequences on the physical dVRK surgical robot. Each pair of sequences shares the same task context, but utilize two different goal shapes (green/red points) sampled from \diffdef{}, resulting in two distinct ways the tissue is retracted.}
    \label{fig:sequence_retraction_tool_real}
\end{figure}

Next, we evaluate \diffdef{} on a real dVRK robotic system manipulating \textit{ex vivo} chicken muscle tissue using real human demonstrations.  
As in surgery, retraction must be coordinated with the cutting tool. 
We model this by introducing a second dVRK instrument that provides context: the tool’s pose dictates which retraction directions are clinically feasible.
Conditioning on this context ensures that predictions respect surgical constraints and proves that complicated surgical contexts can be used to inform \diffdef{}, resulting in multiple unique valid goals.

\paragraph{Dataset} We collected 42 human expert demonstrations across 21 tool poses (two goal shapes per pose).  
We train on 32 demonstrations from 16 tool contexts and hold out the remaining 10 demonstrations from 5 unseen contexts for testing. 

\paragraph{Evaluation} For each test case, we sample one goal point cloud from \diffdef{}. To assess point cloud quality, we compare the generated goal to the two human-demonstrated goals and report the minimum Chamfer distance, since diffusion does not allow us to explicitly choose between modes. A binary success is recorded if the tissue is retracted past the target plane without tool collision.

\paragraph{Results} 

\begin{figure}
    \centering
    \includegraphics[width=1.0\linewidth]{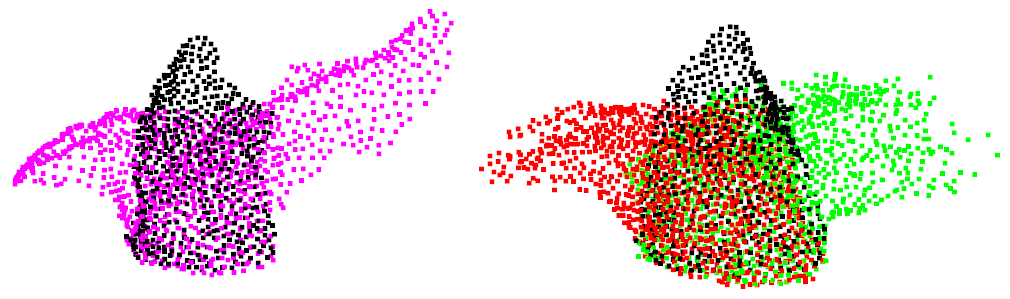} 
    \caption{Worst-case predicted goals from \diffdef{} (right) vs. ground-truth (left) in physical retraction. Black: initial tissue state; Purple: two ground-truth goals; Red/green: two predicted goals. Even in the worst case, predictions closely resemble expert demonstrations.}
    \label{fig:diffdef_vs_gt_real_retraction_tool}
\end{figure}

Across five unseen contexts, chamfer distances were  $0.029$, $0.015$, $0.007$, $0.005$, and $0.009$ millimeters.
Figure~\ref{fig:diffdef_vs_gt_real_retraction_tool} visualizes the \textit{worst-case} predicted goal point cloud among the test set. 
Qualitatively, the goals look similar to the ground truth, and the model captures well the goal shape geometry, showcasing \ggn{}'s ability to generalize even when trained on a relatively small dataset. 

We perform the retraction procedure on three distinct tool pose configurations. 
For each configuration, we execute the full seroving pipeline using \DeformerNet{} on two different goals generated by \diffdef{}.
For each goal predicted by \diffdef{} we run the servoing pipeline three times to assess the robustness of our method, resulting in $3\times2\times3 = 18$ trials in total. 
Across these trials, we observe a retraction success rate of 100\%. 
6 representative sequences are shown in Figure~\ref{fig:sequence_retraction_tool_real}.  

\subsection{Object Packaging}
\label{sec:exp_packaging}
  
In manufacturing, the deformable object packaging task involves safely deforming and compressing an object so that it fits into a container. This scenario is common in warehouse operations, where the deformable item to be packaged may initially be larger than the container’s opening.
This task highlights continuous multimodal distributions, as an infinite number of feasible packaging arrangements exist.

\subsubsection{Simulated Packaging}
\label{sec:exp_packaging_sim}

We set up an object packaging task in Isaac Gym where a deformable pillow must be placed inside a box with varying poses.
In this setting, the partial point cloud of the box serves as contextual input.
Unlike retraction, goal shape non-determinism here arises from continuous rotations of the pillow around the vertical axis. This results in a complex goal distribution characterized by a single high-variance mode, rather than multiple distinct modes.

\begin{figure}
\centering
    \includegraphics[width=0.8\linewidth]{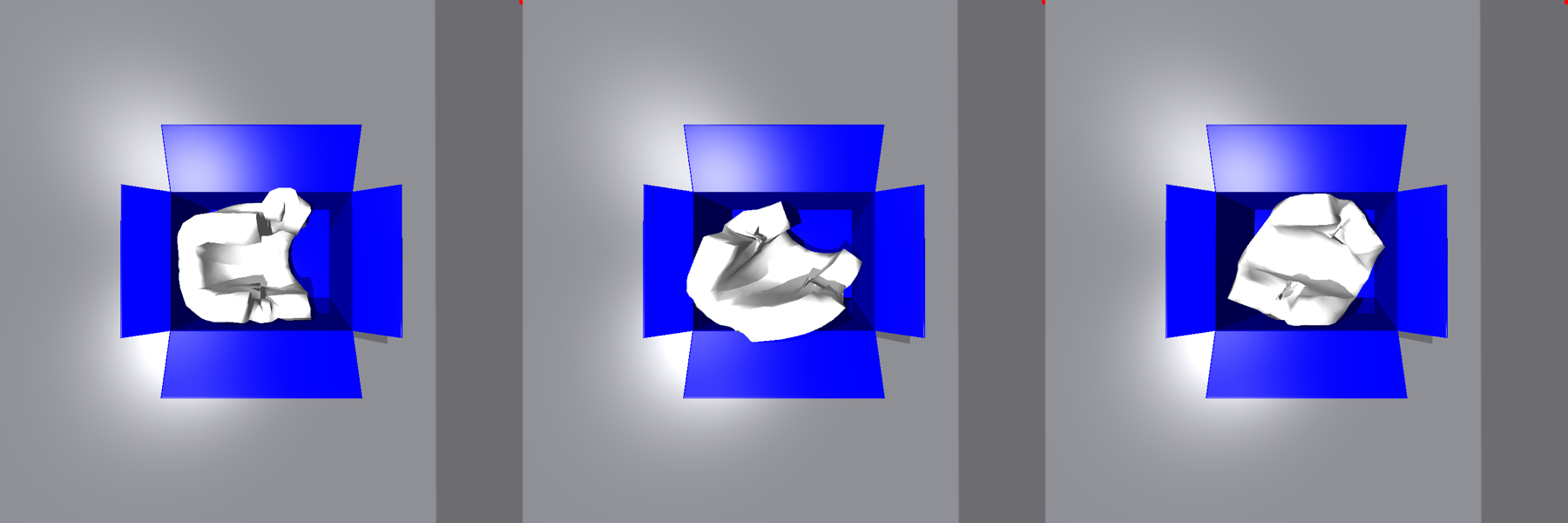}
    \vspace{3pt}
    \caption{Example demonstrations for the simulated object packaging task. For a given container, the deformable object can be arranged in multiple valid configurations.}
    \label{fig:demo_example_packaging}
\end{figure}

\paragraph{Dataset} 
Only for large-scale simulation experiments, demonstrations are collected using a scripted multimodal robot policy (Figure~\ref{fig:demo_example_packaging}). For physical robot experiments, we use real human demonstrations (Figure~\ref{fig:example_demo_real_packaging}). The pillow rotation angle is sampled from a continuous uniform distribution over $(-\pi,\pi]$, yielding a complex continuous goal distribution characterized by a single high-variance mode.

\begin{figure}
    \centering
    \vspace{3pt}
    \includegraphics[width=1\linewidth, trim=250pt 100pt 250pt 100pt, clip]{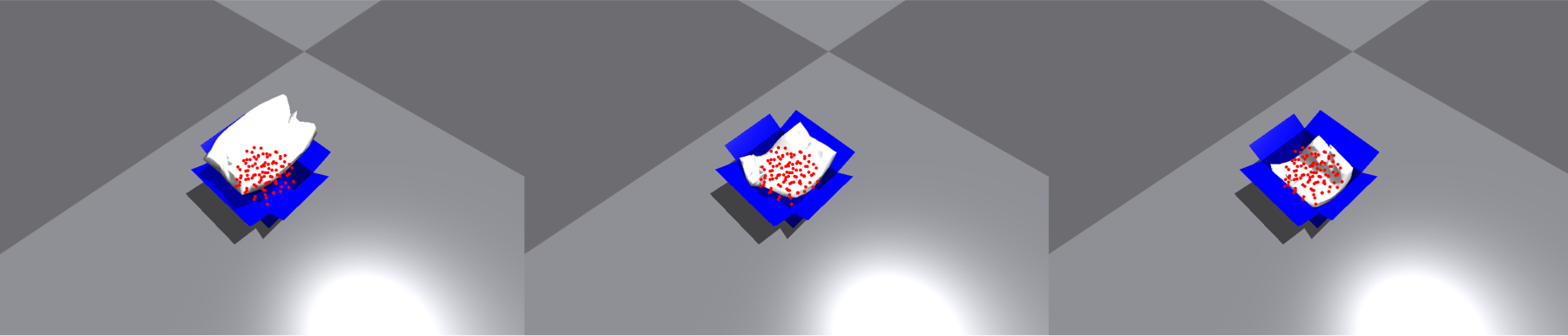}
    \vspace{-9pt}
    \caption{Sample manipulation sequence on the object packaging task, in simulation. The red point cloud visualizes the goal shape generated by \diffdef{}.} 
    \label{fig:sequence_packaging_sim}
\end{figure}

\paragraph{Training and Evaluation}
We train \diffdef{} with 10, 100, and 1000 demonstrations and compare it against \ggn{}, which is trained with 1000 demonstrations. A representative manipulation sequence is shown in Figure~\ref{fig:sequence_packaging_sim}.
The primary evaluation metric is \textit{coverage percentage}, defined as the fraction of object volume contained within the box after servoing with \DeformerNet{}. We additionally report the Chamfer distance between the final object shape and human expert goals.
For each test scenario, we randomize both the deformable object size and the container pose. We then execute the \DeformerNet{} policy conditioned on the goal predicted by \diffdef{}, record the resulting object point cloud, and compute the percentage of points successfully contained.

\begin{figure}
    \centering
    \vspace{5pt}
    \includegraphics[width=1.0\linewidth]{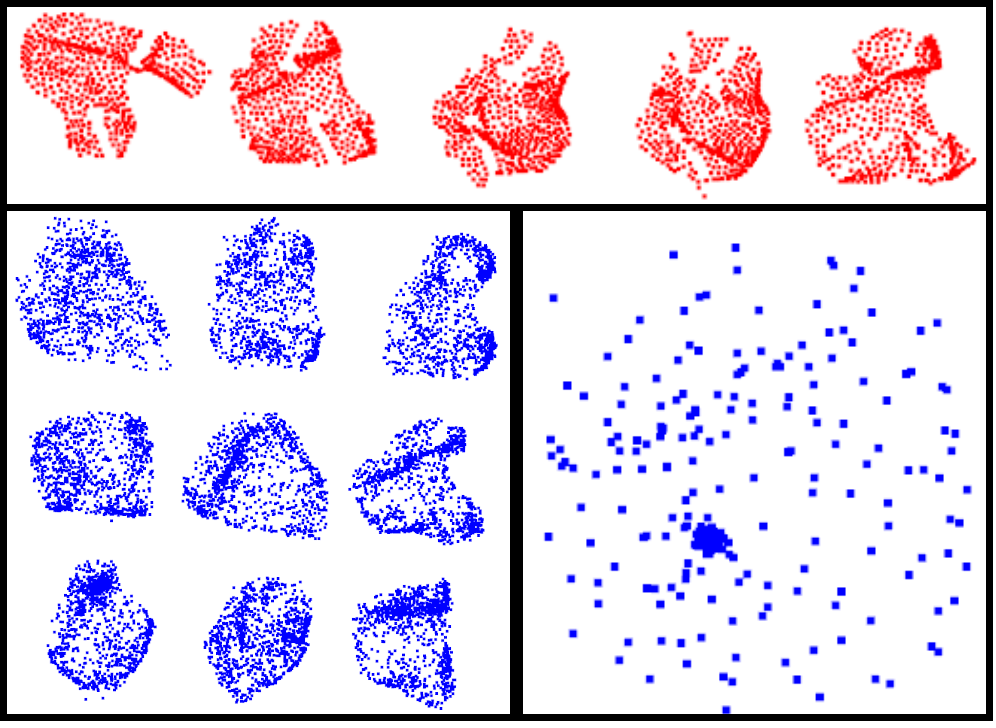}
    \vspace{-9pt}
    \caption{\textbf{(Red)} Ground-truth goal point clouds collected from expert demonstrator. \textbf{(Blue)} Predicted goals from \diffdef{} (bottom left) compared to \ggn{} (bottom right). \diffdef{} generates diverse, multimodal goals, while \ggn{}'s prediction is physically infeasible.}
    \label{fig:diffdef_vs_defgoalnet_packaging}
\end{figure}
\begin{figure}
    \centering
    \vspace{5pt}
    \includegraphics[width=\linewidth]{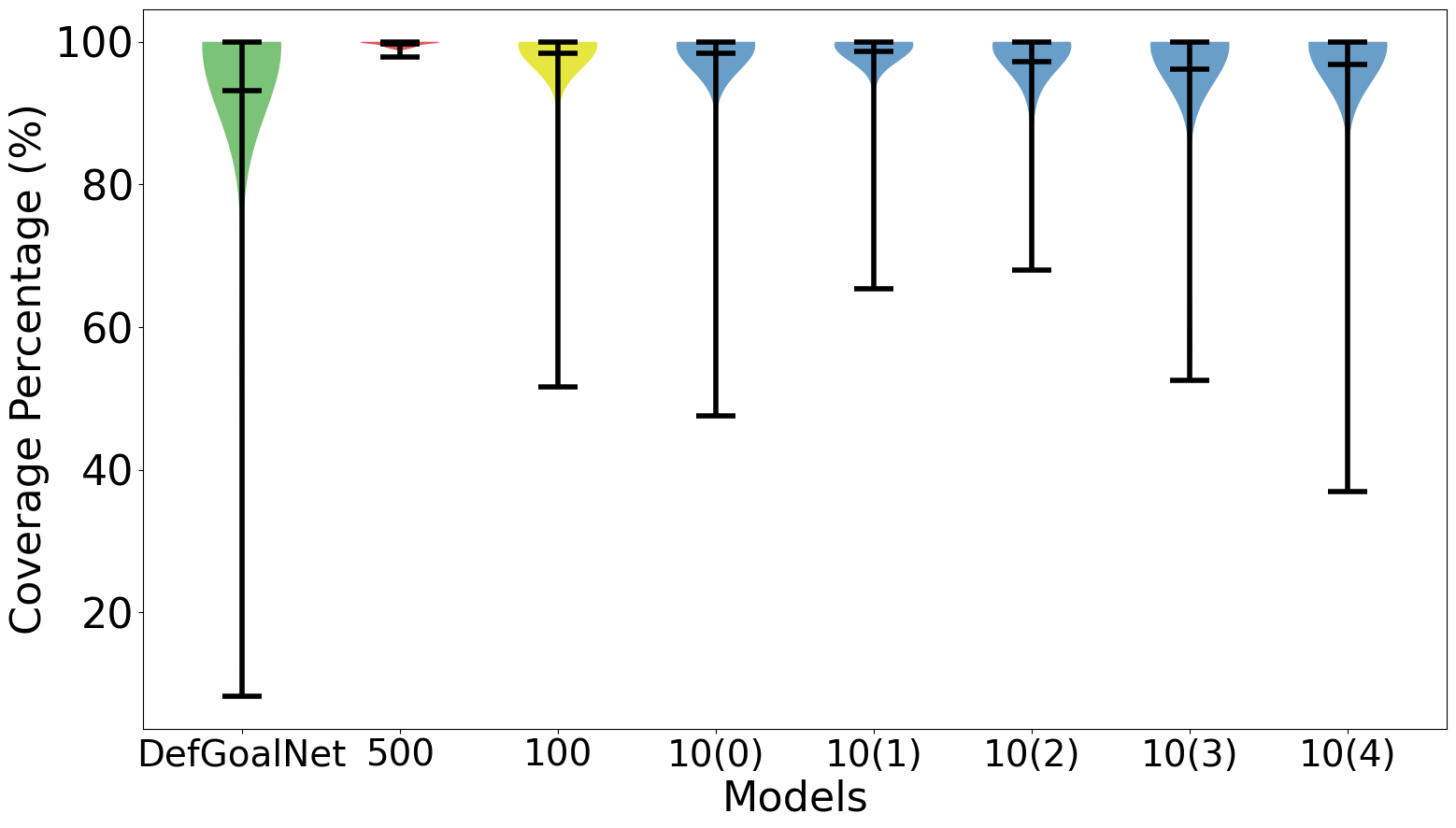}  
    \caption{\textbf{Simulated object packaging}---Violin plots of the coverage percentage metric (higher is better). Left to right: \ggn{}, followed by \diffdef{} trained with 500, 100, and 10 demos.}     \label{fig:object_packaging_eval_coverage_diffdef_vs_defgoalnet}
\end{figure}
\begin{figure}
    \includegraphics[width=\linewidth]{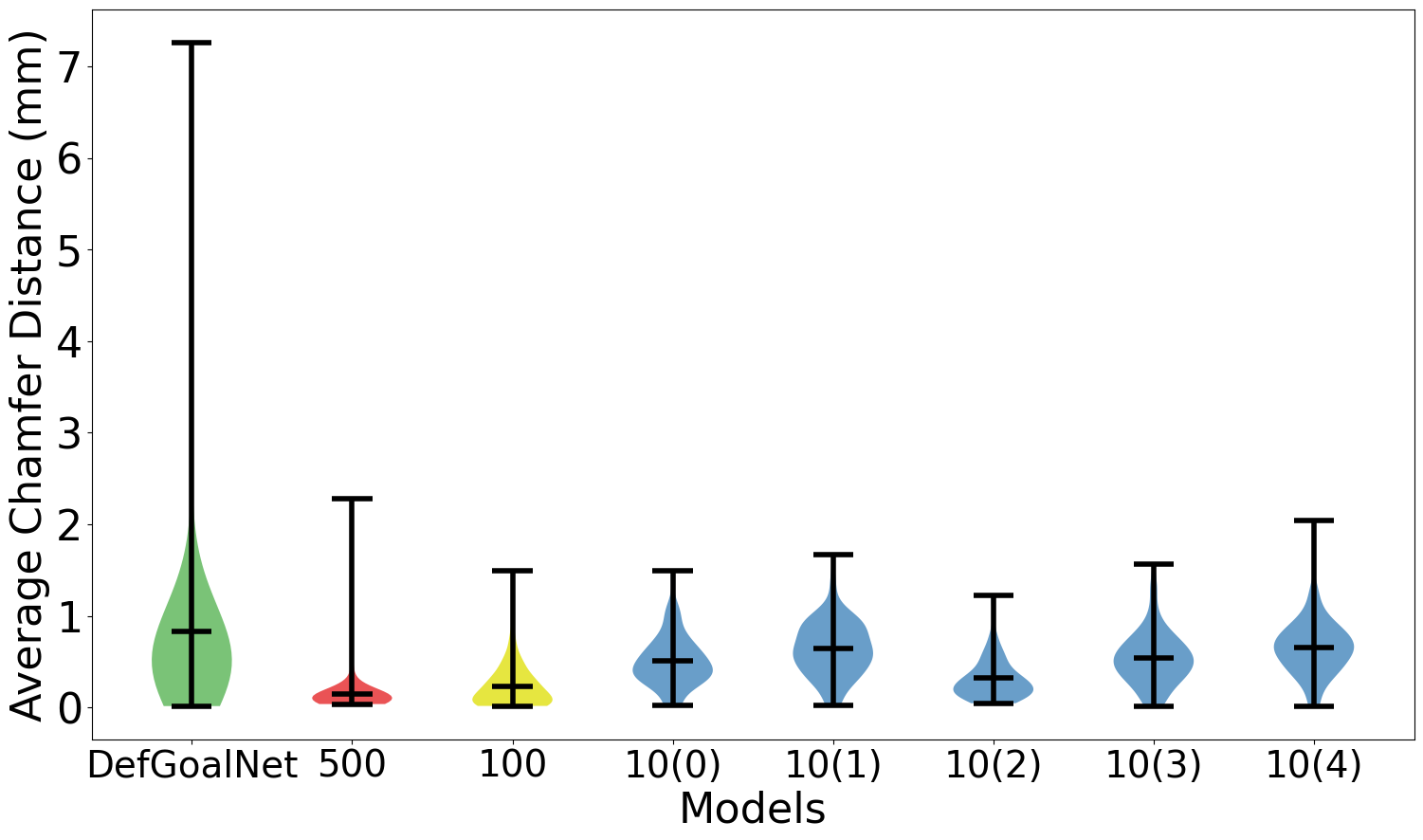}  
     \caption{\textbf{Simulated object packaging}---Violin plots of Chamfer distances (lower is better) between predicted and ground truth goals.}    \label{fig:object_packaging_eval_chamfer_diffdef_vs_defgoalnet}
\end{figure}

\paragraph{Results} 
We present qualitative results in Figure~\ref{fig:diffdef_vs_defgoalnet_packaging}, and quantitative results in Figure~\ref{fig:object_packaging_eval_coverage_diffdef_vs_defgoalnet} and Figure~\ref{fig:object_packaging_eval_chamfer_diffdef_vs_defgoalnet}.
\diffdef{} achieves nearly 100\% median coverage with only 100 demonstrations, outperforming \ggn{} trained with 1000 demonstrations.
\diffdef{} again captures effectively the complex continuous goal distribution, while \ggn{}  produces physically impractical goal point clouds.
Figure~\ref{fig:object_packaging_eval_chamfer_diffdef_vs_defgoalnet} shows the Chamfer distance between predicted and ground-truth goal point clouds on a test set of 100 unseen demonstrations. Similar to the surgical retraction task, \diffdef{} significantly outperforms \ggn{}.

\subsubsection{Physical Object Packaging}
\label{sec:exp_packaging_real}

Finally, we validate \diffdef{} on physical hardware using the dual-arm KUKA iiwa system (Figure~\ref{fig:physical_exp_setup}-left). Two 7-DOF arms equipped with Robotiq and Reflex Takktile 2 grippers are tasked with placing a soft pillow into a cardboard box. 

\begin{figure}
    \centering
    \includegraphics[width=0.8\linewidth]{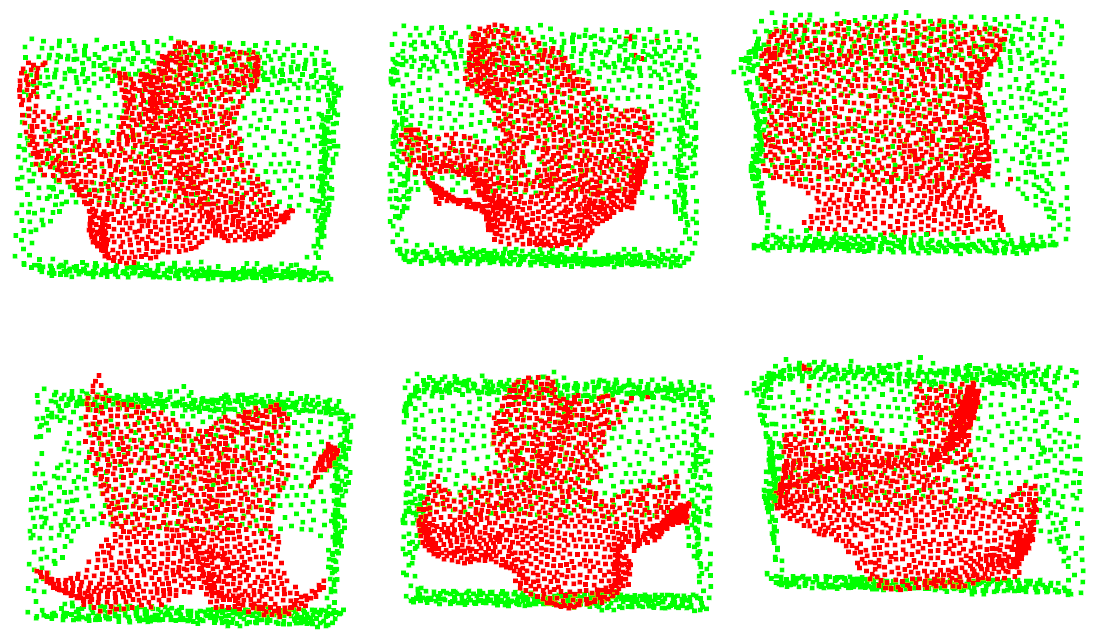} 
    \caption{Example human demonstrations for real-robot object packaging task. Green: container (context); Red: six equally valid goal shapes representing different pillow placement strategies.}
    \label{fig:example_demo_real_packaging}
\end{figure}

\paragraph{Dataset} We collected 36 demonstrations across six distinct box poses, with six valid pillow placements for each pose (Figure~\ref{fig:example_demo_real_packaging}).
We use 30 demonstrations from five box poses for training and hold out the remaining six demonstrations from the sixth pose for testing. 

\begin{figure}
    \centering
    \includegraphics[width=0.8\linewidth]{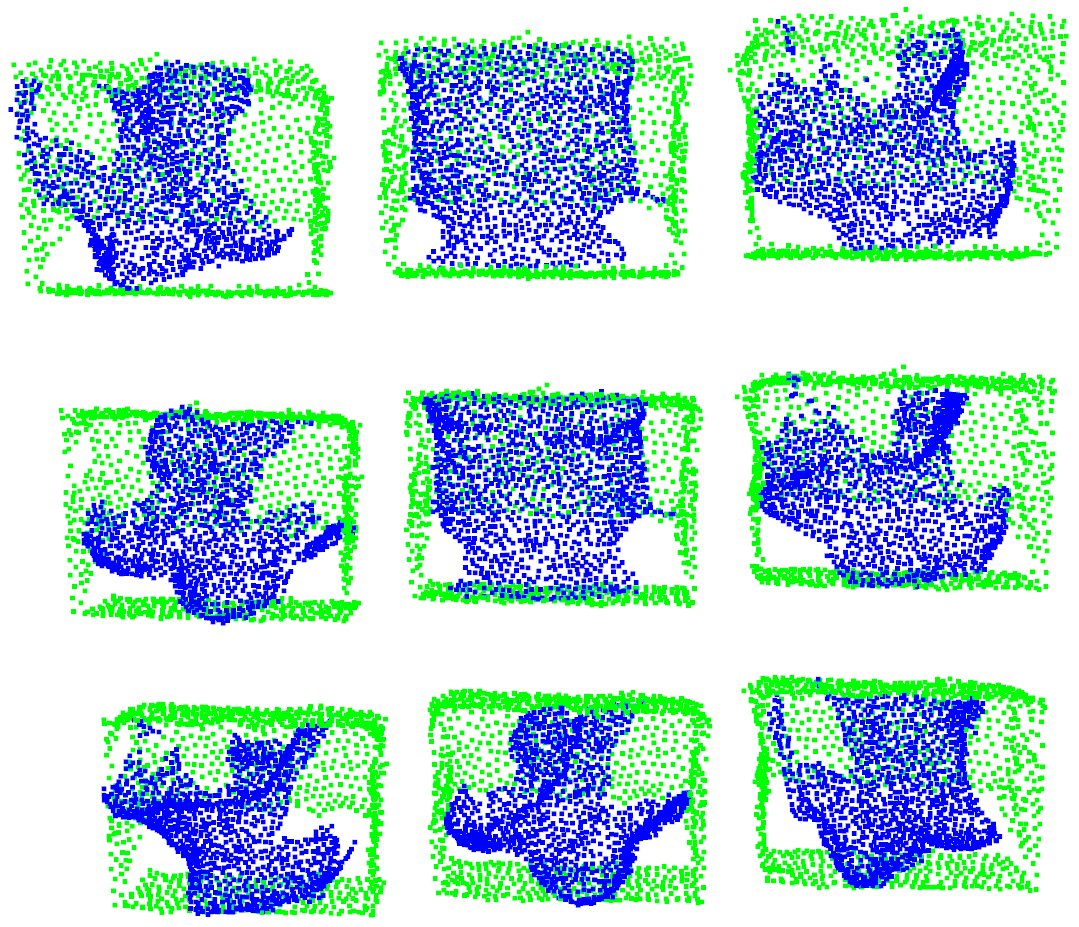} 
    \caption{Predicted goal point clouds (blue) sampled from the goal distribution learned by \diffdef{}, in real-robot packaging experiment.}
    \label{fig:diffdef_vs_gt_real_packaging}
\end{figure}

\begin{figure}
    \centering
    \includegraphics[width=\linewidth]{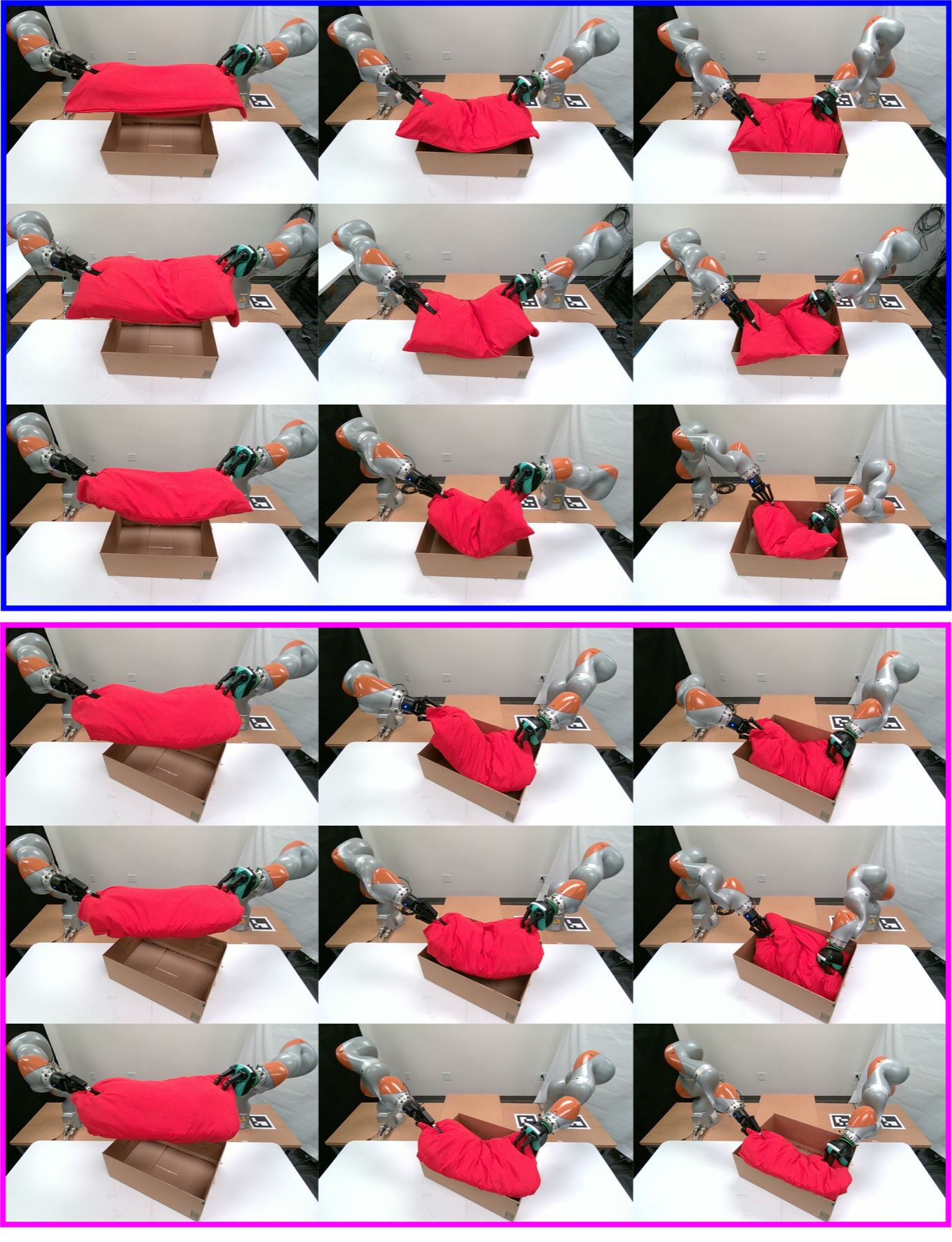}  
    \caption{Six object packaging sequences with the KUKA system. Each context (container's pose) supports multiple valid packaging strategies corresponding to distinct goals predicted by \diffdef{}.}
    \label{fig:sequence_packaging_real}
\end{figure}

\paragraph{Results} On the held-out box pose, the predicted goal shapes closely matched the human demonstrations (Figure~\ref{fig:diffdef_vs_gt_real_packaging}).
We evaluated the full servoing pipeline (\diffdef{} + \DeformerNet{}) on two previously unseen box poses, sampling three predicted goals per pose and conducting three trials for each goal ($2 \times 3 \times 3 = 18$ trials in total).
All 18 trials were successful, with representative sequences shown in Figure~\ref{fig:sequence_packaging_real}.

\subsection{Summary of Results}
  
Across all tasks and domains, \diffdef{} consistently generates diverse and realistic goal shapes that closely resemble human demonstrations, leading to high task success rates.  
The surgical retraction task highlights context-aware multimodality under safety-critical constraints in medical settings, while the object packaging task illustrates continuous multimodality in industrial manipulation.  
Together, these results demonstrate that \diffdef{} is a general and versatile framework for deformable object manipulation, well-suited for diverse real-world conditions and application domains.

\section{Conclusion}\label{sec:conclusions}
We have presented a pipeline for learning deformable object manipulation from demonstrations. 
At the heart of this pipeline is \diffdef{}, a neural network that learns a multimodal distribution over diverse goal shapes capable of successfully completing the given task.
We evaluate \diffdef{} across a broad range of robotic tasks spanning both surgery and manufacturing applications. Our experiments show that \diffdef{} consistently outperforms the current state-of-the-art goal prediction method---which models a single deterministic goal rather than a distribution as in our approach---across multiple metrics and task domains, while also requiring less training data.
Notably, our approach enables effective multimodal task-specific goal generation from only a limited number of demonstrations, while still benefiting from a generic control policy trained on a large and diverse dataset that is cheap to acquire and agnostic to the specific downstream task.

% \section*{ACKNOWLEDGMENT}
% This work was supported in part by NSF Awards \#2024778 and \#2133027.

%%%%%%%%%%%%%%%%%%%%%%%%%%%%%%%%%%%%%%%%%%%%%%%%%%%%%%%%%%%%%%%%%%%%%%%%%%%%%%%%

% \clearpage\newpage
\bibliographystyle{IEEEtran}
\bibliography{references}

\end{document}